\documentclass{article} 
\usepackage{iclr2026_conference,times}

\usepackage{amsmath,amsfonts,bm}

\def\eqref#1{equation~\ref{#1}}

\def\1{\bm{1}}

\DeclareMathAlphabet{\mathsfit}{\encodingdefault}{\sfdefault}{m}{sl}
\SetMathAlphabet{\mathsfit}{bold}{\encodingdefault}{\sfdefault}{bx}{n}

\usepackage{hyperref}
\usepackage{url}
\usepackage{booktabs}
\usepackage{graphicx}
\usepackage{subcaption}
\usepackage{multirow}
\usepackage{enumitem}
\usepackage{wrapfig}
\usepackage{caption}

\title{Multi-Step Reasoning for Embodied Question Answering via Tool Augmentation}
\iclrfinaltrue

\author{Mingliang Zhai$^{1,2}$\thanks{Equal contribution.}~~, 
Hansheng Liang$^{3*}$, 
Xiaomeng Fan$^{1*}$, 
Zhi Gao$^{1}$, \\
\textbf{Chuanhao Li}$^{4}$,
\textbf{Che Sun}$^{2}$, 
\textbf{Xu Bin}$^{3}$, 
\textbf{Yuwei Wu}$^{1,2}$, 
\textbf{Yunde Jia}$^{2}$ \\
\small $^1$ Beijing Key Laboratory of Intelligent Information Technology, Beijing Institute of Technology, Beijing, China \\
\small $^2$ Guangdong Laboratory of Machine Perception and Intelligent Computing, Shenzhen MSU-BIT University, \\Shenzhen, Guangdong, China \\
\small $^3$ School of Mechanical Engineering, Beijing Institute of Technology, Beijing, China \\
\small $^4$ Shanghai Artificial Intelligence Laboratory, Shanghai, China
}

\begin{document}

\maketitle
\vspace{-3em}
\begin{figure}[h]
    \includegraphics[width=\textwidth]{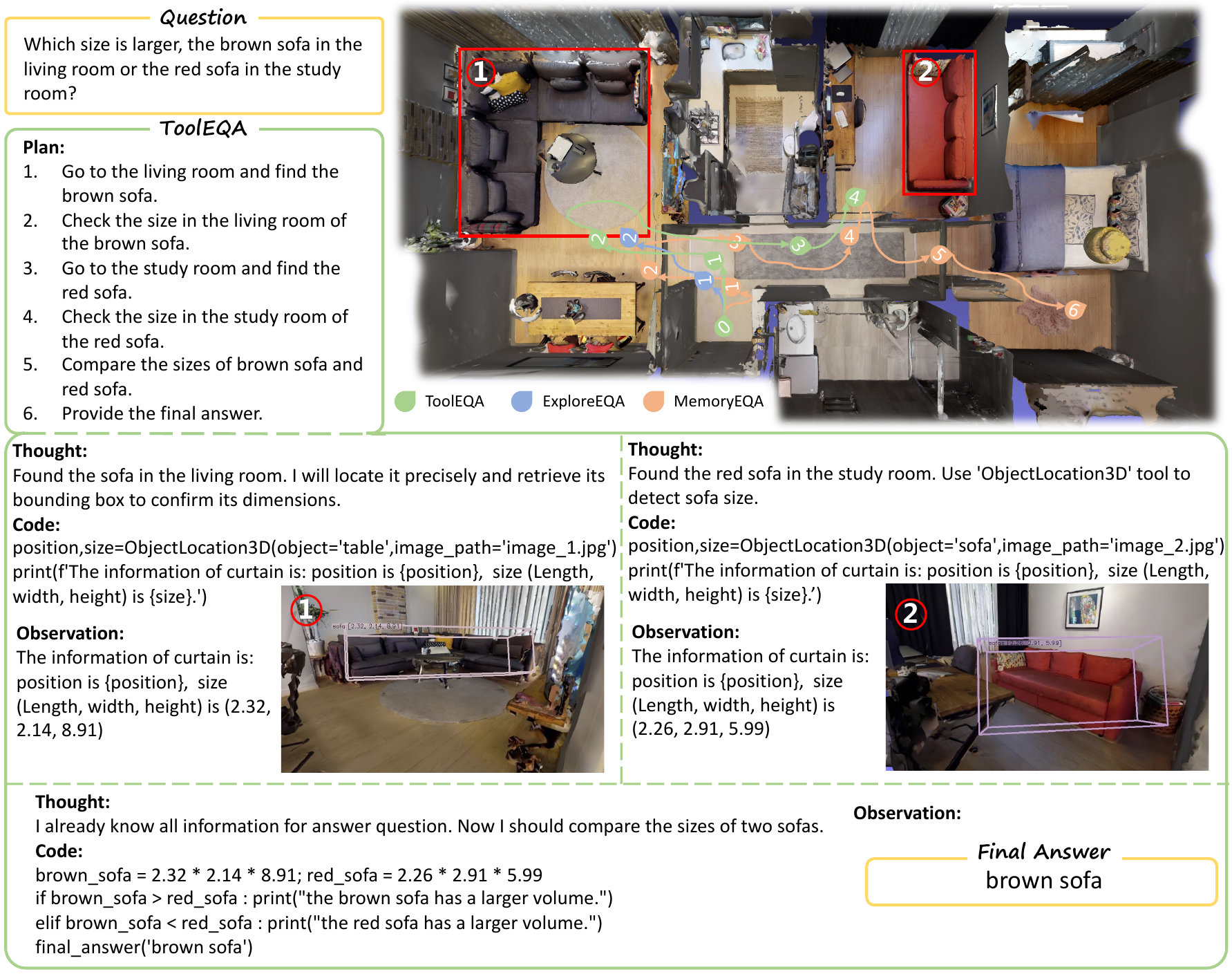}
    \vspace{-2em}
    \caption{
    Overview of the proposed ToolEQA for Embodied Question Answering (EQA). ToolEQA enables to decompose questions into structured plans, reasoning to select tools, and invoke tools to explore and answer.  ToolEQA achieves highest accuracy with fewer reasoning steps.
    }
    \label{fig:case}
\end{figure}
\vspace{-0.5em}

\begin{abstract}
\vspace{-1em}
Embodied Question Answering (EQA) requires agents to explore 3D environments to obtain observations and answer questions related to the scene.
Existing methods leverage VLMs to directly explore the environment and answer questions without explicit thinking or planning, which limits their reasoning ability and results in excessive or inefficient exploration as well as ineffective responses.
In this paper, we introduce \textbf{ToolEQA}, an agent that integrates external tools with multi-step reasoning, where external tools can provide more useful information for completing the task, helping the model derive better exploration directions in the next step of reasoning and thus obtaining additional effective information. 
This enables ToolEQA to generate more accurate responses with a shorter exploration distance.
To enhance the model’s ability for tool-usage and multi-step reasoning, we further design a novel EQA data generation pipeline that automatically constructs large-scale EQA tasks with reasoning trajectories and corresponding answers.
Based on the pipeline, we collect the EQA-RT dataset that contains about 18K tasks, divided into a training set EQA-RT-Train, and two test sets EQA-RT-Seen (scenes overlapping with the training set) and EQA-RT-Unseen (novel scenes).
Experiments on EQA-RT-Seen and EQA-RT-Unseen show that ToolEQA improves the success rate by 9.2$\sim$20.2\% over state-of-the-art baselines, while outperforming the zero-shot ToolEQA by 10\% in success rate. 
In addition, ToolEQA also achieves state-of-the-art performance on the HM-EQA, OpenEQA, and EXPRESS-Bench datasets, demonstrating its generality.
Our homepage see \url{https://tooleqa.github.io}.

\end{abstract}

\vspace{-1em}
\section{Introduction}
\vspace{-1em}

Embodied Question Answering (EQA), a challenging task in computer vision and robotics, requires agents to navigate in a 3D environment, actively gather visual information through exploration, and answer questions about the scene \citep{das2018embodied}. 
Existing methods \citep{ziliotto2025tango, ren2024explore, zhai2025memory, cheng2024efficienteqa,jiang2025beyond} leverage VLMs to understand environment for guiding exploration and answering questions, but they generally lack explicit intermediate reasoning and planning.
For example, as shown in Figure~\ref{fig:case}, 
(1) The agent often answers the question before fully identifying all relevant objects, resulting in incorrect final answers due to insufficient information gathering capabilities.
(2) The agent makes suboptimal route plans, prolonging the exploration process and reduce efficiency due to limited reasoning abilities.
This motivates us to leverage tools to enhance the information-gathering capabilities of the agent, and use multi-step explicit reasoning to improve its reasoning ability during the exploration process, enabling it to complete EQA tasks with more efficient exploration distances.

In this paper, we propose ToolEQA, an agent that leverages tool augmentation to perform multi-step reasoning for EQA tasks. ToolEQA reasons over both current observations and historical information, selects appropriate tools to invoke, and integrates the additional information they provide (e.g., 3D bounding boxes) into the reasoning process. To ground reasoning in the environment, we abstract the action space into tool sets and execute them as actions. The agent iteratively reasons and applies tools, acquiring new observations until the final answer is derived. By effectively integrating collected information and identifying shorter exploration paths, ToolEQA improves both exploration efficiency and accuracy in solving EQA tasks.

To enhance the reasoning capability of the ToolEQA agent, we introduce a novel EQA data generation pipeline that automatically generates large-scale EQA tasks with reasoning trajectories via three steps: EQA task generation, reasoning trajectory generation, and validation. 
Specifically, we first employ a 3D detection model to identify all objects in the current scene and extract their attributes, such as size and spatial coordinates. 
Based on this object-level information, we then leverage GPT-4o~\citep{gpt4o} to automatically generate diverse questions and their corresponding answers. 
Subsequently, to generate optimal reasoning trajectories, we extract all relevant objects mentioned in the question and determine the shortest path by combining their positions with an A-star algorithm. 
On top of this path, we incorporate reasoning steps and tool usage into the path by employing GPT-4o to generate complete trajectories.
To ensure the correctness of questions, we design question-type-specific prompt templates that guide the generation process, thereby ensuring both path optimality and consistency in task solving.
Finally, to preserve data quality, the generated EQA tasks and trajectories are passed through an EQA task verifier and trajectory verifier to discard low-quality data and rectify incorrect trajectories.

With the data generation pipeline, we construct EQA-RT, a dataset of 18K EQA question–answer pairs with reasoning trajectories. 
We further split it into a training set (EQA-RT-Train) and two test sets, where two test sets contain EQA-RT-Seen (in-domain scenes overlapping with the training set) and EQA-RT-Unseen (out-of-domain scenes for evaluating generalization). 
We train the proposed ToolEQA agent on EQA-RT-Train using supervised fine-tuning.
We comprehensively evaluate the tuned ToolEQA agent and the zero-shot ToolEQA agent on HM-EQA~\citep{ren2024explore}, Open-EQA~\citep{majumdar2024openeqa}, ExpressBench~\citep{jiang2025beyond}, EQA-RT-Seen and EQA-RT-Unseen.
The ToolEQA agent consistently achieves improvements on untrained VLMs and outperforms them by 11\%. 
This indicates that our method enables agents to have powerful capability for practical EQA tasks with complex and diverse trajectories.
In summary, our contributions are three-fold.
\vspace{-1em}
\begin{itemize}[leftmargin=*]
    \item We propose the ToolEQA agent which performs multi-step reasoning for environment exploration and question answering, achieving improved effectiveness and efficiency in solving EQA tasks.
    

    \item We introduce an EQA data generation pipeline that automatically generates large-scale EQA tasks with reasoning trajectories. 
    
    \item We introduce EQA-RT, a dataset containing 18K question–answer pairs for EQA, covering diverse and complex question types with high-quality reasoning trajectories.
    
\end{itemize}

\vspace{-1em}
\section{Realted Works}
\vspace{-1em}
\subsection{Embodied question answering}
\vspace{-0.5em}
Embodied question answering \citep{das2018embodied, gordon2018iqa, yu2019multi, cangea2019videonavqa,das2018neural} has become a challenging paradigm for testing a robot's ability to autonomously plan tasks and establish semantic understanding of the environment in order to correctly answer natural language questions. 
\cite{yu2019multi} constructed  a multi-target question answering dataset in a virtual environment and introduced a multi-target EQA  method.
\cite{ren2024explore} first applied VLMs to EQA and built the HM-EQA dataset with more open-ended questions for realistic and diverse evaluation. Subsequent studies extended VLMs for EQA. \cite{majumdar2024openeqa} used video memory for implicit questions with long-context VLMs.  \cite{saxena2024grapheqa} embedded a planner into compact scene representations to bridge semantic memory and planning. 
\cite{jiang2025beyond} added exploration-trajectory annotations to EQA datasets and incorporated exploration into evaluation metrics.
However, those methods demonstrate limited reasoning capacity, as the lack of explicit thinking and planning, which often leads to redundant or inefficient exploration.
To solve this problem, we introduce a multi-step reasoning process for solving EQA task via tool augmentation, enhancing effectiveness and efficiency.

\vspace{-1em}
\subsection{Multi-step Reasoning}
\vspace{-0.5em}
Multi-step reasoning can significantly enhance a model’s ability to solve complex tasks while improving interpretability. 
In recent years, research on multi-step reasoning in large language models (LLMs)~\citep{ranaldi2024empowering,openai2024o1,chen2024m,yao2025mmreason} and multi-modal systems has made notable progress. 
\cite{li2024vocot} proposed the VoCoT framework, which integrates vision-guided and object-centric chain-of-thought reasoning to improve the reasoning performance of large-scale multimodal models on complex tasks. 
The ReAct~\citep{yao2023react} framework is a multi-step reasoning paradigm that decomposes tasks through iterative Reason–Act–Observe cycles, which greatly benefits to solving challenging problems. 
In the domain of Embodied Question Answering (EQA), Fine-EQA~\citep{jiang2025beyond} introduced a new benchmark that emphasizes dynamic exploration and multi-step reasoning in 3D environments, aiming to improve both exploration efficiency and evaluation metrics. 
However, the exploration of multi-step reasoning in embodied question answering remains at a relatively early stage.
We propose ToolEQA, which integrates explicit multi-step reasoning into the multi-step exploration process in embodied scenarios. This step-by-step thinking strategy not only shortens exploration paths but also enhances the accuracy of question answering.

\vspace{-1em}
\subsection{Tool Usage Agent}
\vspace{-0.5em}
Recent work have equipped VLMs with tool-usage capabilities. 
Frameworks such as ReAct \citep{yao2023react} and Toolformer \citep{schick2023toolformer} demonstrated the effectiveness of coupling reasoning traces with tool execution, while embodied agents like SayCan \citep{ahn2022can} showed how language-guided tool usage can translate high-level instructions into low-level actions. 
T3-Agent \citep{gao2024multi} leveraged automatically generated multimodal tool-usage data and fine-tunes vision-language models (VLMs) as controllers to enable strong tool-based reasoning.
~\cite{li2025adaptive} proposed the MeCo framework, which captures the model’s ``cognitive signals'' to assess its capability boundaries and thereby decide whether to invoke external tools. 
These works enhance multi-modal reasoning by calling predefined tools to acquire additional information. 
However, such methods perform reasoning only within static cyberspace. In contrast, we define the physical environment itself as a tool, thereby situating reasoning steps within embodied interactions and enabling more autonomous embodied agents.

\begin{figure}[t]
    \centering
    \includegraphics[width=0.9\linewidth]{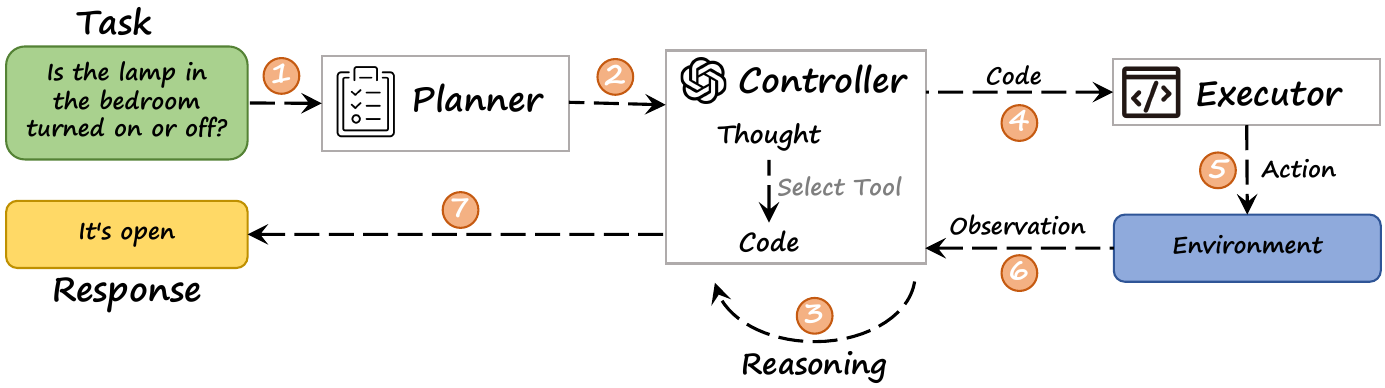}
    \caption{Overview of the ToolEQA agent workflow. 
    }
    \label{fig:agent}
    \vspace{-1.5em}
\end{figure}

\vspace{-1em}
\section{ToolEQA Agent}
\vspace{-1em}




To enable the agent to reason and act in complex environments, we propose a ToolEQA agent that integrates tool-usage strategies for the reasoning process. 
ToolEQA conducts step-by-step reasoning on past observations and, at each step, generates corresponding thoughts and code to execute tools.
Code offers greater flexibility than formats such as JSON for handling diverse inputs and outputs.
As shown in Figure~\ref{fig:agent}, ToolEQA comprises three components: a planner for generating overall task plan $p$, a controller for generating thought $t$ and code $c$, and an executor for executing code in environment. 
Given a query $Q$ and a scene $S$, the $i$-step of the agent is formulated as
\begin{equation}
    \begin{aligned}
        t_{i}^{*} , c_{i}^{*} = \arg \max P(t_i, c_i|Q, S, h_{i}, p),
    \end{aligned}
\end{equation}
where $t_{i}^{*}$ and $c_{i}^{*}$ are generated thought and code for the $i$-th step, and $h_{i} = \{t_{1}, c_{1}, o_{1}, ... , t_{i-1}, c_{i-1}, o_{i-1}\}$ is the history (thought, code, and observation of previous steps).

\vspace{-1em}
\paragraph{Planner.}
Given an EQA task, the planner, modeled as an LLM, takes the query as input, interprets the task objectives, and outputs an overall plan that decomposes the task into sub-goals.
The structured sub-goals are provided to the controller
to prevent blind exploration, enhancing the efficiency and accuracy of task execution.

\vspace{-1em}
\paragraph{Executor.} 
We deploy real-executable tools for the agent. Our tools include \texttt{GoNextPoint}, \texttt{ObjectLocation2D}, \texttt{ObjectLocation3D}, \texttt{ObjectCrop}, \texttt{VisualQA}, \texttt{FinalAnswer}, the details of tools see Appendix~\ref{sec:tools_desc}.
With the generated code, the executor calls executable tools in the environment to obtain new observations for further exploration, thus solving the EQA task.

\vspace{-1em}
\paragraph{Controller.} 
The controller performs dynamic reasoning for deciding which tool to use and executing it, on the basis of the question and the previous observations, with the guidance of plans. By invoking the executor to gather new observations for further reasoning until the answer is derived. 
The reasoning process can be divided into three situations. 
\vspace{-0.5em}
\begin{itemize}[leftmargin=*]
    
    \item \textbf{The collected information is insufficient and the current scene lacks required objects.} ToolEQA infers missing objects from previous observations and the query, and estimates their likely positions. Then, ToolEQA combines these estimates with its current location to decide a walking direction, and uses the `\texttt{GoNextPoint}' (for example, `\texttt{GoNextPoint}(``turn left'')') to gather the needed information.


    \item \textbf{The collected information is insufficient and the current scene contains involved objects.} ToolEQA reasons over the question and invokes suitable tools to obtain relevant information. For example, as shown in Figure~\ref{fig:case}, ToolEQA utilizes `\texttt{ObjectLocation3D}' to extract the size of objects, and then continues exploration following the above step.

    \item \textbf{The collected information is sufficient.} ToolEQA needs to reason over the question, use appropriate tools on the current image, and integrate information for the final answer. For example, it processes the detected object, then writes `Python' codes for comparing sizes, and employs `\texttt{FinalAnswer}' to produce the ultimate output.

\end{itemize}








\begin{figure}[t]
    \centering
    \includegraphics[width=\linewidth]{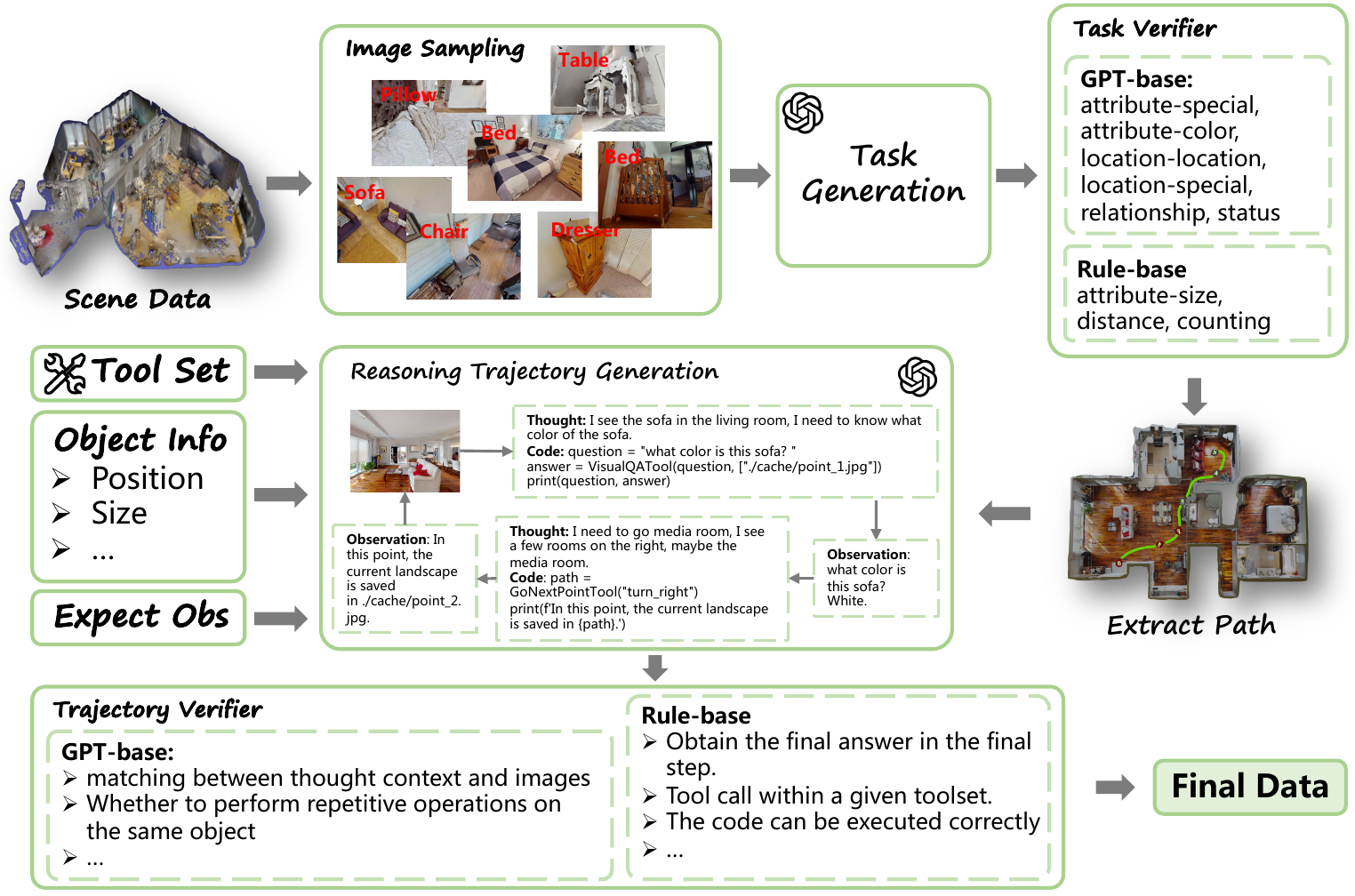}
    \caption{ EQA Data Generation Pipeline.
    }
    \vspace{-1.5em}
    \label{fig:pipeline}
\end{figure}

\vspace{-1em}
\section{EQA Data Generation Pipeline}
\vspace{-1em}
\subsection{Formulation}
\vspace{-0.5em}
 
\paragraph{Data Format.} 
We format the EQA tool-usage data as $\{S, pos, Q, p, T, C, O, A\}$, where $S$ denotes the scene, $pos$ denotes the initial position of agents, $Q$ denotes the question, $p$ denotes the overall plan for solving the task, $T$ denotes the generated thought, $C$ denotes the generated code, $O$ denotes observation (outputs of using tools), and $A$ means the ground truth answer.
Considering that solving one real-world EQA task may require multiple steps involving multiple tools, $T$, $C$, and $O$ can be represented by the integration of thought, code, and observation in multiple steps, and the data format is reformulated as $\{S, pos, Q, p, \{t_{1}, \cdots, t_{n}\}, \{c_{1}, \cdots, c_{n}\}, \{o_{1}, \cdots, o_{n}\},A\}$, where $t_{i}, c_{i}$, and $o_{i}$ indicate the thought, code, and observation in the $i$-th step respectively, and there are $n$ steps in total. The thought, code, and observation are composed of a trajectory $\{t_{1}, c_{1}, o_{1}, \cdots, t_{n}, c_{n}, o_{n}\}$ of $n$ steps to solve the task. 
\vspace{-0.5em}
\paragraph{Scene Source.} 
HM3D \citep{ramakrishnan2021habitat} is a comprehensive dataset comprising 3D reconstructions of 1,000 large-scale buildings collected from diverse real-world locations. 
We select 713 high-quality scenes from HM3D as our data source, sample object images from them, and generate questions and answers. 
\vspace{-0.5em}

The proposed data generation pipeline is shown in Figure~\ref{fig:pipeline}, including three steps: EQA task generation, reasoning trajectory generation, and data verification.

\vspace{-1em}
\subsection{EQA task generation}
\vspace{-0.5em}

Our goal is to generate a large set of diverse, practical, and complex EQA tasks. 
We first apply a 3D detection model to obtain each object’s bounding box, position, and category, and sample the object image from detected objects.
The object attributes and corresponding visual information are then fed into GPT-4o along with example question-answer pairs designed from brainstorming to simulate natural home conversations. 
Guided by the prompt, GPT-4o generates questions and answers across six types: relationship, status, distance, location, counting, and attribute, where location is divided into two subcategories `location-location' and `location-special', and attribute is divided into three subcategories `color', `special', and `size'. 
The answers are open-ended or multiple-choice, enabling the evaluating different capabilities of agents.



\vspace{-1em}
\subsection{Reasoning Trajectory Generation}
\vspace{-0.5em}
Given an EQA task, we construct an exploration trajectory that records reasoning steps, tool selections, and observations. The trajectory is constrained to follow the shortest path and ensure consistency between reasoning and tool usage. We extract objects mentioned in the question  using their locations and the agent’s position, and compute the shortest path using the A$^*$ algorithm, generating intermediate waypoints and navigation directions.

Based on these trajectories, GPT-4o enriches each step with reasoning and tool selections. 
Steps are categorized as key, where the target object is found, and non-key, where it is not. 
For non-key steps, GPT-4o receives the current image and exploration direction to generate reasoning. 
For the key steps, we select possible tools required to solve the task from the toolset, and then prompt GPT-4o to output which specific tool should be invoked under the current observation and the corresponding rationale.
To ensure consistency and rationality, we design question-type–specific prompts containing task-specific considerations, reasoning strategies, tool-usage guidelines, and examples, allowing GPT-4o to produce thought and code across different question types.

\vspace{-1em}
\subsection{Data Verification}
\vspace{-0.5em}
To preserve the quality of generated data, we design an EQA task verifier and a trajectory verifier to filter out low-quality data. 
Using LLMs to verify generated tasks and trajectories has proven effective \citep{gao2024multi, liu2024apigen}. Inspired by this, we use LLMs to verify generated tasks and trajectories.


\noindent \textbf{EQA Task Verifier.} Since object descriptions in generated questions or options may not always match the scene, we use two complementary strategies: confidence-based matching and LLM-based structured scoring to evaluate quality and filter out low-quality samples. 
For confidence, we first extract object descriptions from the question and options, locate the corresponding objects in the scene to obtain their images, and then use Grounded-SAM~\citep{ren2024grounding} to compute a score reflecting how well each image matches its description.
For LLM-based scoring, we feed the question and object images into GPT-4o, which outputs a similarity score. 
We set thresholds for the two strategies respectively, and samples below thresholds on either score are filtered out.


\noindent \textbf{Trajectory Verifier.}
To verify the rationality of tool usage and reasoning in the generated trajectories, we adopt two strategies: rule-based and LLM-based validation.
For the rule-based validation, we design several checks: 
(1) the key tools should exist and be invoked at the correct step (\textit{e.g.}, \texttt{GoNextPoint} should be called at every step before reaching the target); 
(2) the invoked tool should belong to the predefined tool set, and its parameters should be passed correctly. 
For the LLM-based validation, we prompt GPT-4o to consider the following factors: 
(1) the predicted answer should be semantically consistent with the ground-truth answer; 
(2) the reasoning in non-key steps should avoid hallucinations; 
(3) the final answer should be reasonably derived from the reasoning and observations.
We set a threshold for three checks, and samples below the threshold are filtered out.

\vspace{-1em}
\subsection{EQA-RT}
\vspace{-0.5em}

\begin{figure}[h]
    \centering
    \begin{subfigure}[b]{0.3\textwidth}
        \centering
        \includegraphics[width=\textwidth]{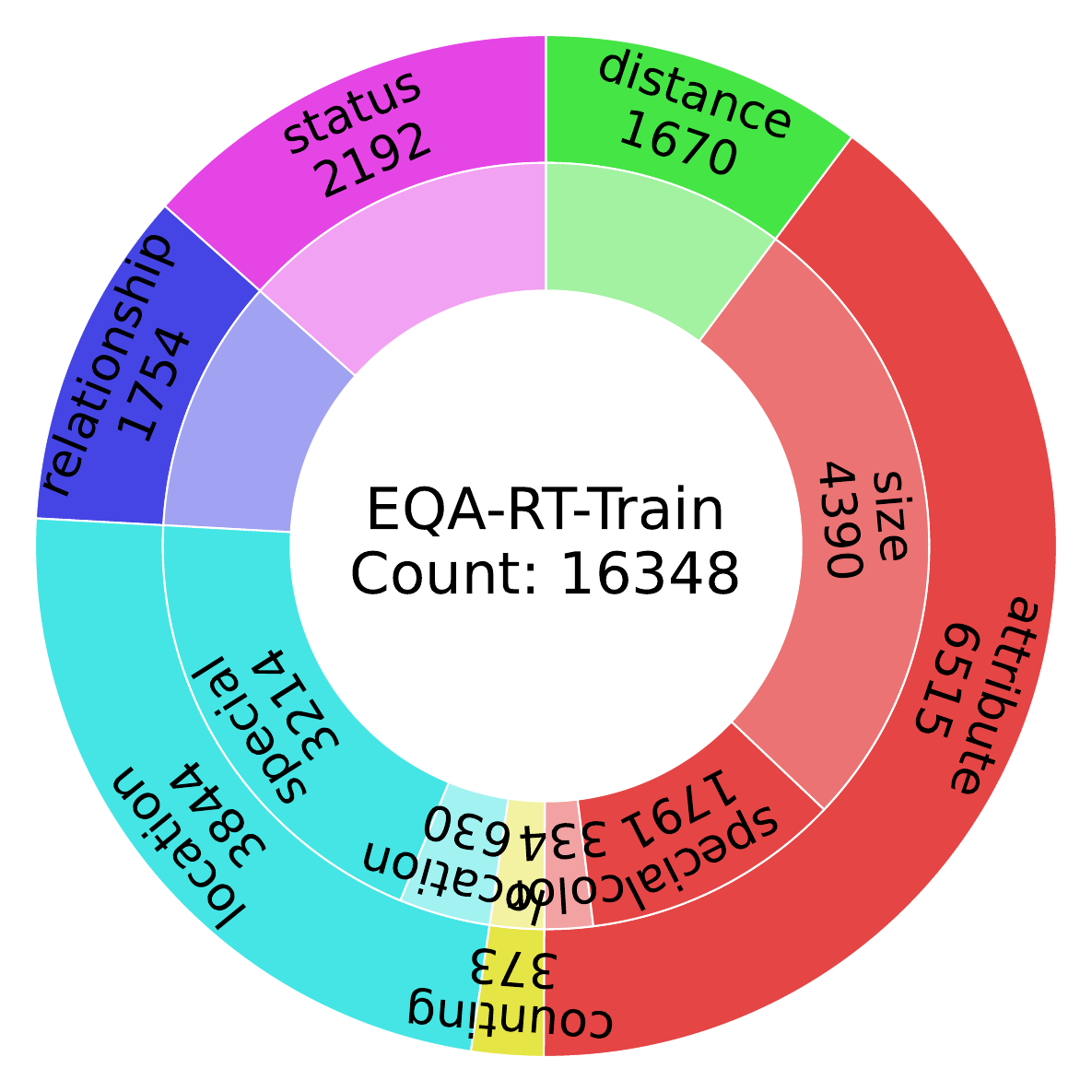}
        \caption{EQA-RT-Train}
        \label{fig:trainset}
    \end{subfigure}
    \hfill
    \begin{subfigure}[b]{0.3\textwidth}
        \centering
        \includegraphics[width=\textwidth]{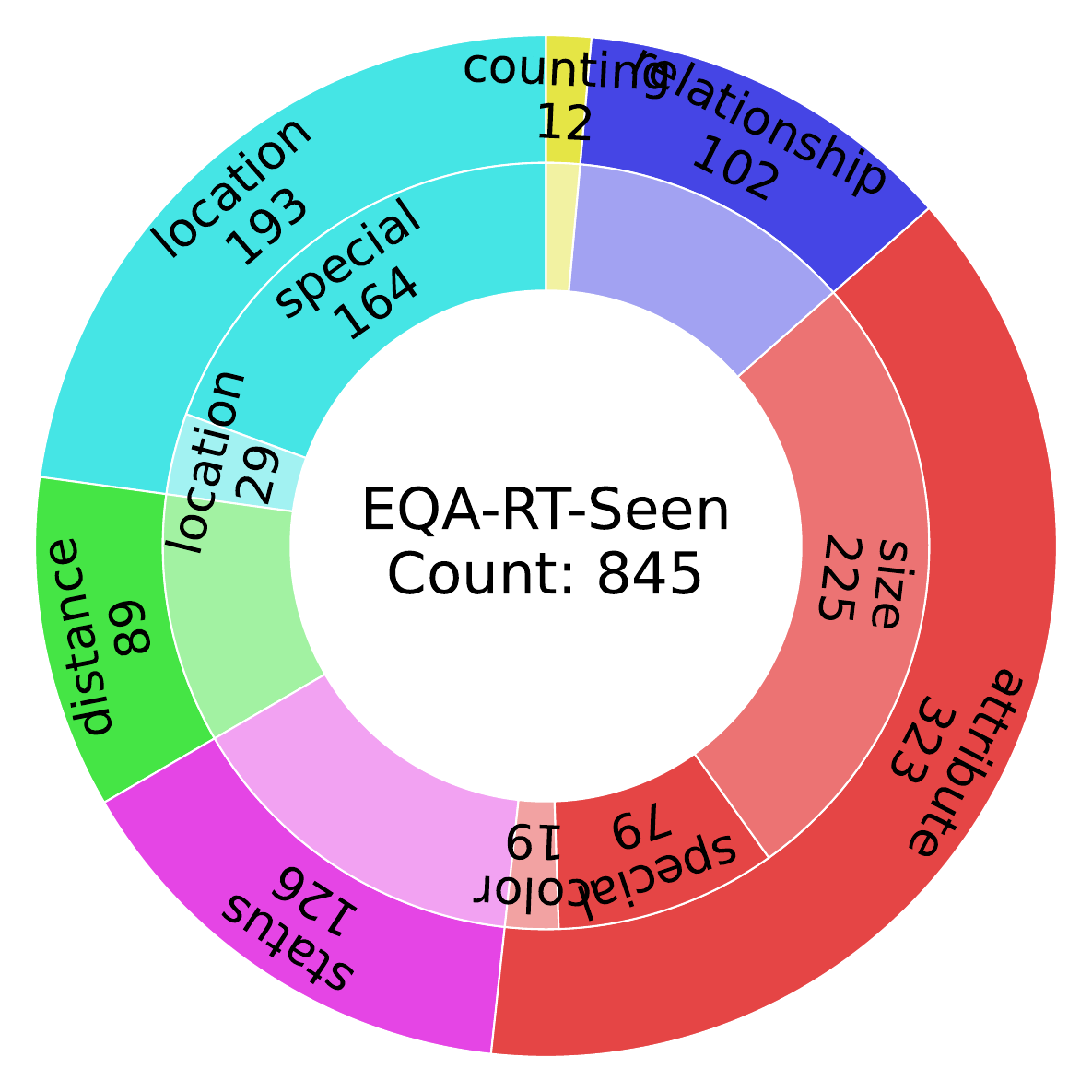}
        \caption{EQA-RT-Seen}
        \label{fig:trainset}
    \end{subfigure}
    \hfill
    \begin{subfigure}[b]{0.3\textwidth}
        \centering
        \includegraphics[width=\textwidth]{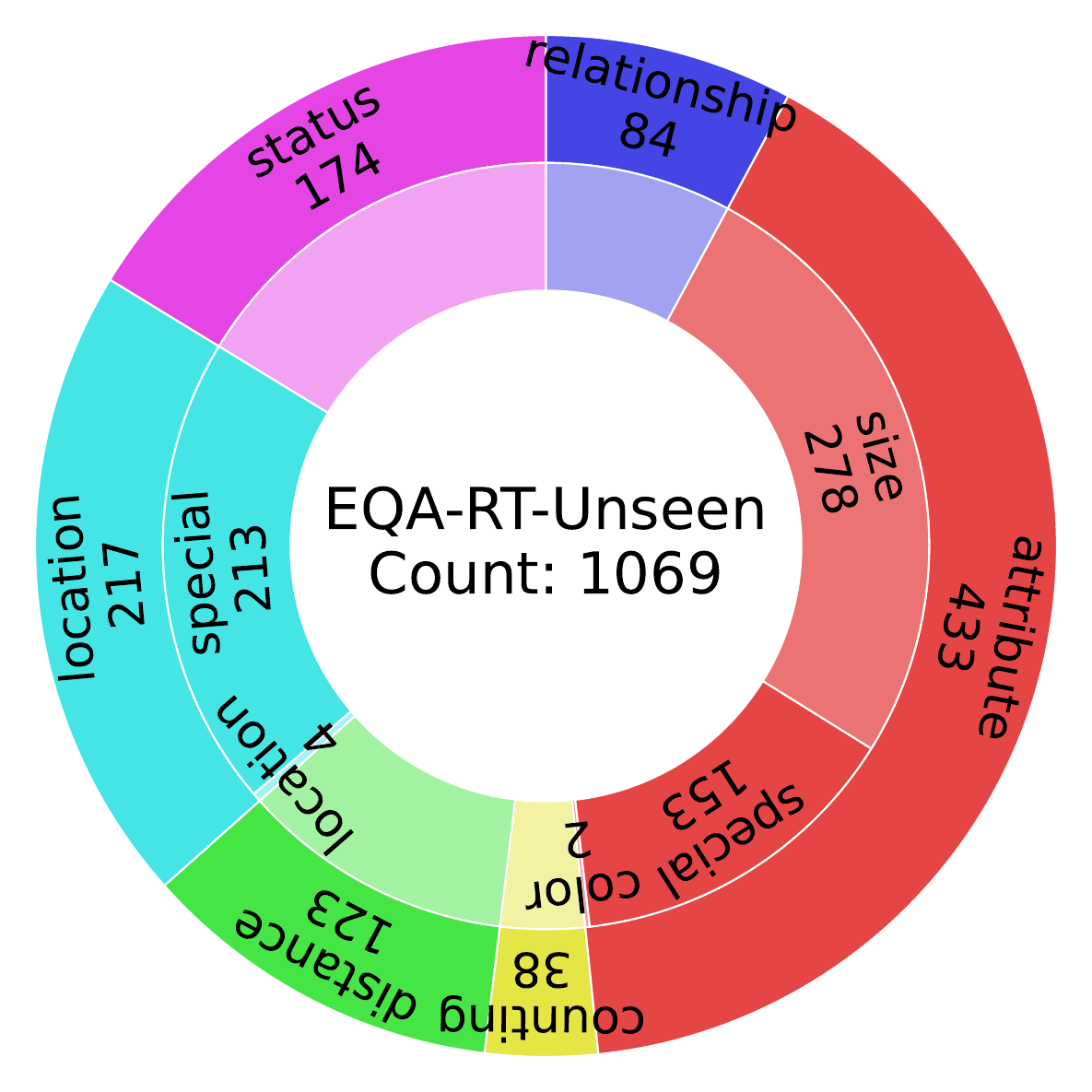}
        \caption{EQA-RT-Unseen}
        \label{fig:testset}
    \end{subfigure}
    
    \caption{Data statistics of the training set (EQA-RT-Train) and two test sets (EQA-RT-Seen and EQA-RT-Unseen). The scenes in EQA-RT-Seen have the overlap with EQA-RT-Train, while the scenes in EQA-RT-Unseen are not present in the training set.}
    \label{fig:statistic}
\end{figure}
By utilizing the developed EQA task generation pipeline, we construct EQA-RT, encompassing about 18K EQA tasks. We further split it into a training set (EQA-RT-Train), a seen test set (EQA-RT-Seen) and a unseen test set (EQA-RT-Unseen), where the test set contains both in-domain scenes overlapping with the training set and out-of-domain scenes for evaluating generalization. 
It is worth noting that the tasks in EQA-RT-Seen are more challenging than those in EQA-RT-Unseen, with detailed evidence provided in Appendix~\ref{sec:data_statistic}.
As shown in Figure~\ref{fig:statistic}, we show the question types of the generated EQA tasks in EQA-RT-train, EQA-RT-Seen and EQA-RT-Unseen. More statistical data can be found in Appendix~\ref{sec:data_statistic}.



\vspace{-1em}
\subsection{Training}
\vspace{-0.5em}
Given a data point $\{S, pos, Q, p \{t_{1}, ..., t_{n}\}, \{c_{1}, ..., c_{n}\}, \{o_{1}, ..., o_{n}\}, A\}$, we train the VLM controller using the cross-entropy loss,
\vspace{-1em}
\begin{equation}
    \begin{aligned}
        \min \mathbb{E}_{(Q,S,pos,T,C,O,A)\sim \mathbb{D}}  \left [ -\sum_{i=1}^{n} P(t_{i}, c_{i}|Q, S, pos, h_{i}) \right ],
    \end{aligned}
\end{equation}
where $\mathbb{D}$ is the EQA-RT-Train dataset and we sum the loss values of the n steps in the trajectory. 
Note that, in training VLMs, we do not fit the final answer $A$, as we encourage the controller to leverage tools in solving given tasks, instead of directly producing an answer based on biases in VLMs. 
The average length of exploration and reasoning trajectories reaches 12.69 steps (as shown in the statistics in Appendix Table~\ref{tab:step_tool_length}). A longer number of steps results in an extended trajectory history $h$, which in turn enlarges the model input and ultimately causes substantial time and memory consumption during training. 
To address this issue, we propose a trajectory sampling strategy. Specifically, we retain the key steps and randomly sample an equal number of non-key steps to reduce resource overhead. This design is motivated by the fact that non-key steps dominate the exploration process and are highly redundant, and they mostly consist of repeated direction predictions and frequent use of the \texttt{GoNextPoint} tool. In contrast, key steps involve diverse tool usage and reasoning changes.
After training, ToolEQA agent can present powerful ability of reasoning and tool-usage, further enhance the effectiveness and efficiency of solving EQA tasks.



\begin{table}[t]
    \centering
    \caption{Baseline evaluation on EQA-RT-Seen.}
    \vspace{-0.5em}
    \small
    \begin{tabular}{c|c|ccc|ccc|c}
    \toprule
        \multirow{2}{*}{Setting} & \multirow{2}{*}{Model} & \multicolumn{3}{c|}{$recall$ $\uparrow$} & \multicolumn{3}{c|}{$e_{path}$ $\uparrow$} & \multirow{2}{*}{succ. (\%) $\uparrow$} \\
        & & $@5$ & $@10$ & $@15$ & $@5$ & $@10$ & $@15$ & \\
    \midrule
        \multirow{5}{*}{\shortstack{Multi \\ Choices}} 
        & Explore-EQA  & 0.06 & 0.11 & 0.14 & 0.04 & 0.07 & 0.09 & 44.7 \\
        & Memory-EQA  & 0.06 & 0.12 & 0.13 & 0.04 & 0.07 & 0.11 & 48.2 \\
        & ToolEQA (gpt-4o)   & 0.06 & 0.14 & 0.19 & \textbf{0.08} & 0.2 & 0.27 & 55.37 \\
        & ToolEQA (qwen2.5vl)  & 0.04 & 0.09 & 0.11 & 0.06 & 0.13 & 0.17 & 53.1 \\
        & ToolEQA (qwen2.5vl ft)  & \textbf{0.06} & \textbf{0.15} & \textbf{0.21} & 0.07 & \textbf{0.23} & \textbf{0.3} & \textbf{57.31} \\
    \midrule
        \multirow{5}{*}{\shortstack{Open \\ Vocabulary}} 
        & Explore-EQA  & 0.04 & 0.10 & 0.13 & 0.04 & 0.06 & 0.09 & 30.6 \\
        & Memory-EQA  & 0.05 & 0.10 & 0.13 & 0.04 & 0.09 & 0.11 & 35.1 \\
        & ToolEQA (gpt-4o)   & 0.06 & 0.15 & \textbf{0.21} & 0.07 & 0.2 & 0.27 & 49.2 \\
        & ToolEQA (qwen2.5vl)  & 0.05 & 0.12 & 0.16 & 0.03 & 0.10 & 0.14 & 44.9 \\
        & ToolEQA (qwen2.5vl ft) & \textbf{0.06} & \textbf{0.15} & 0.20 & \textbf{0.08} & \textbf{0.22} & \textbf{0.3} & \textbf{53.6} \\
    \bottomrule
    \end{tabular}
    \label{tab:seen}
    \vspace{-0.5em}
\end{table}
\begin{table}[t]
    \centering
    \caption{Baseline evaluation on EQA-RT-Unseen.}
    \vspace{-0.5em}
    \small
    \begin{tabular}{c|c|ccc|ccc|c}
    \toprule
        \multirow{2}{*}{Setting} & \multirow{2}{*}{Model} & \multicolumn{3}{c|}{$recall$ $\uparrow$} & \multicolumn{3}{c|}{$e_{path}$ $\uparrow$} & \multirow{2}{*}{succ. (\%) $\uparrow$} \\
        & & $@5$ & $@10$ & $@15$ & $@5$ & $@10$ & $@15$ & \\
    \midrule
        \multirow{5}{*}{\shortstack{Multi \\ Choices}} 
        & Explore-EQA  & 0.06 & 0.12 & 0.15 & 0.05 & 0.08 & 0.10 & 47.0 \\
        & Memory-EQA  & 0.06 & 0.13 & 0.14 & 0.06 & 0.09 & 0.11 & 48.9 \\
        & Too l (gpt-4o)   & 0.07 & \textbf{0.16} & \textbf{0.21} & 0.08 & 0.2 & 0.28 & 57.9 \\
        & ToolEQA (qwen2.5vl)  & 0.04 & 0.11 & 0.13 & 0.07 & 0.16 & 0.21 & 55.3 \\
        & ToolEQA (qwen2.5vl ft)  & \textbf{0.07} & 0.14 & 0.19 & \textbf{0.08} & \textbf{0.24} & \textbf{0.3} & \textbf{59.5} \\
    \midrule
        \multirow{5}{*}{\shortstack{Open \\ Vocabulary}} 
        & Explore-EQA  & 0.05 & 0.09 & 0.15 & 0.05 & 0.10 & 0.13 & 31.4 \\
        & Memory-EQA & 0.05 & 0.09 & 0.15 & 0.06 & 0.15 & 0.18 & 35.9 \\
        & ToolEQA (gpt-4o)   & 0.06 & 0.16 & 0.21 & 0.06 & 0.21 & 0.27 & 49.3 \\
        & ToolEQA (qwen2.5vl)  & 0.06 & 0.13 & 0.17 & 0.05 & 0.16 & 0.2 & 45.1 \\
        & ToolEQA (qwen2.5vl ft)   & \textbf{0.08} & \textbf{0.17} & \textbf{0.24} & \textbf{0.09} & \textbf{0.24} & \textbf{0.32} & \textbf{56.1} \\
    \bottomrule
    \end{tabular}
    \label{tab:unseen}
    \vspace{-0.5em}
\end{table}

\vspace{-1em}
\section{Experiments}
\vspace{-1em}

\subsection{Setting}
\vspace{-0.5em}
we tested ToolEQA on the EQA-RT and HM-EQA \citep{ren2024explore} datasets and compared it with existing open-source methods, Explore-EQA \citep{ren2024explore} and Memory-EQA \citep{zhai2025memory}. We also examined the impact of different models (GPT-4o \citep{gpt4o}, Qwen2.5-VL-7B \citep{wang2024qwen2}, and fine-tuned Qwen2.5-VL-7B \citep{wang2024qwen2}) as controllers on performance. In addition, we conducted a qualitative analysis of ToolEQA, investigating how reasoning and tool invocation affect the efficiency and success rate of completing EQA tasks.

\vspace{-0.5em}
\paragraph{Training}
We trained the controller using the EQA-RT training set. 
During the training of the VLM-based controller, we froze the vision encoder and the visual token compressor, and fine-tuned the language model with LoRA \citep{hu2022lora}. 
We adopted the AdamW optimizer with a cosine annealing scheduler, using a learning rate of 1e-6 and a batch size of 1. We used 4 Nvidia Tesla H100 GPUs to train for 2 days.

\vspace{-0.5em}
\paragraph{Metrics}
We use three metrics for evaluating ToolEQA and existing EQA methods.
The success rate is divided into two parts, for multi-choices tasks, we calculate average accuracy between the output of model and ground truth answer; for open vocabulary tasks, we prompt LLM to obtain the semantic similarity between the output of model and ground truth answer.
$recall@D$ is used to evaluate whether objects related to the problem were found during the
exploration process.
$e_{path}@D$ is an indicator that combines success rate, recall, and exploration path length.
The details of the metrics can be found in Appendix~\ref{sec:details_of_metric}.


\begin{wraptable}{r}{0.55\textwidth}
\vspace{-1em}
    \centering
    \caption{Comparison between the original model and the finetuned model in terms of the number of key steps, thought length, correct tool usage rate, and success rate on the EQA-RT-Unseen dataset.}
    \vspace{-1em}
    \small
    \begin{minipage}{0.55\textwidth}
        \centering
        \setlength{\tabcolsep}{4pt}
        \begin{tabular}{c|cccc}
            \toprule
                Model & Step & Thought & Tool & Succ.\\
            \midrule
                ToolEQA (0-shot) & 1.24 & 90.26 & 58 & 45.1 \\
                ToolEQA (ft) & 1.98 & 116.15 & 69 & 56.1 \\
            \bottomrule
        \end{tabular}
        \label{tab:thought}
        \vspace{1em}
    \end{minipage}
    \hfill
    \small
    \begin{minipage}{0.55\textwidth}
        \caption{Performance comparison on EXPRESS-Bench.}
        \vspace{-1em}
        \begin{tabular}{c|cccc}
        \toprule
            Model & Succ. $\uparrow$ & Succ.$^*\uparrow$ & E$_{path}\uparrow$ & d${_T}\downarrow$ \\
        \midrule
            Fine-EQA & 40.55 & 63.95 & 16.22 & 6.43 \\
            ToolEQA & 42.21 & 65.77 & 25.82 & 5.25\\
        \bottomrule
        \end{tabular}
        \label{tab:express-bench}
        \vspace{1em}
    \end{minipage}
    \hfill
    \small
    \begin{minipage}{0.55\textwidth}
        \centering
        \caption{EQA-Agent performance on existed benchmarks. $^\dagger$ represents that the metric comes from the our implementation.}
        \vspace{-1em}
        \begin{tabular}{c|cc|cc}
            \toprule
               \multirow{2}{*}{Model} & \multicolumn{2}{c|}{HM-EQA} & \multicolumn{2}{c}{OpenEQA} \\
                & succ.(\%) & L(m) & succ.(\%) & L(m)\\
            \midrule
                Explore-EQA   & 51.5          & 38.87          & 28.3$^\dagger$ & - \\
                Efficient-EQA & 54.3          & 30.16          & -              & - \\
                Memory-EQA    & 63.4          & 33.54          & 34.6$^\dagger$ & - \\
                Graph-EQA     & \textbf{63.5} & -              & 30.1$^\dagger$ & - \\
                ToolEQA       & 62.3          & \textbf{18.26} & \textbf{35.5}  & \textbf{6.96} \\
            \bottomrule
        \end{tabular}
        \label{tab:benchmark}
        \vspace{-1em}
    \end{minipage}
\end{wraptable}


\vspace{-1em}
\subsection{Main Results}
\vspace{-0.5em}
As shown in Table~\ref{tab:seen} and Table~\ref{tab:unseen}, we report the performance of different methods on EQA-RT-Seen and EQA-RT-Unseen. Our ToolEQA consistently outperforms reasoning-inefficient methods Explore-EQA~\citep{ren2024explore} and Memory-EQA~\citep{zhai2025memory} across all metrics, demonstrating its effectiveness in tackling complex tasks.
The comparison between agents equipped with fine-tuned and non-fine-tuned VLMs further validates the effectiveness of our data generation pipeline.
The success rate of fine-tuned Qwen2.5VL-7B compared to the original Qwen2.5VL-7B on EQA-RT-Unseen improved from 45.1 to 56.1, the recall rate increased from 0.17 to 0.24, and $e_{path}$ improved from 0.2 to 0.32.
Compared with the non-fine-tuned Qwen2.5-VL-7B, ToolEQA with GPT-4o achieves better performance, indicating that the controller’s capability directly determines the performance of ToolEQA. 
However, the fine-tuned Qwen2.5-VL-7B surpasses GPT-4o in $e_{path}$ and success rate, while achieving comparable recall. 
This indicates that our training has enabled the VLM to learn how to think and solve problems more effectively in indoor scenarios.


We typically consider that the length of thoughts is positively correlated with reasoning ability~\citep{jin2024impact}. Therefore, we evaluate the impact of fine-tuning the VLM on EQA-RT-Unseen with respect to thought length and the accuracy of tool calls (i.e., the proportion of calls that successfully acquire the information required to answer the question).
As shown in Table~\ref{tab:thought}, after fine-tuning, thought length increases from 90.26 to 116.15, tool call accuracy improves from 58 to 69, and the success rate rises from 45.1 to 56.1. This indicates that reasoning ability is crucial for accomplishing the EQA task.

As shown in Table~\ref{tab:benchmark}, we compare our method with Fine-EQA~\citep{jiang2025beyond} on EXPRESS-Bench~\citep{jiang2025beyond}, and with Explore-EQA~\citep{ren2024explore}, Efficient-EQA~\citep{cheng2024efficienteqa}, Memory-EQA~\citep{zhai2025memory}, and Graph-EQA~\citep{saxena2024grapheqa} on HM-EQA~\citep{ren2024explore} and OpenEQA~\citep{majumdar2024openeqa}. Our method achieves comparable success rates with shorter exploration distances across all these datasets.

\begin{figure}[t]
    \centering
    \includegraphics[width=\linewidth]{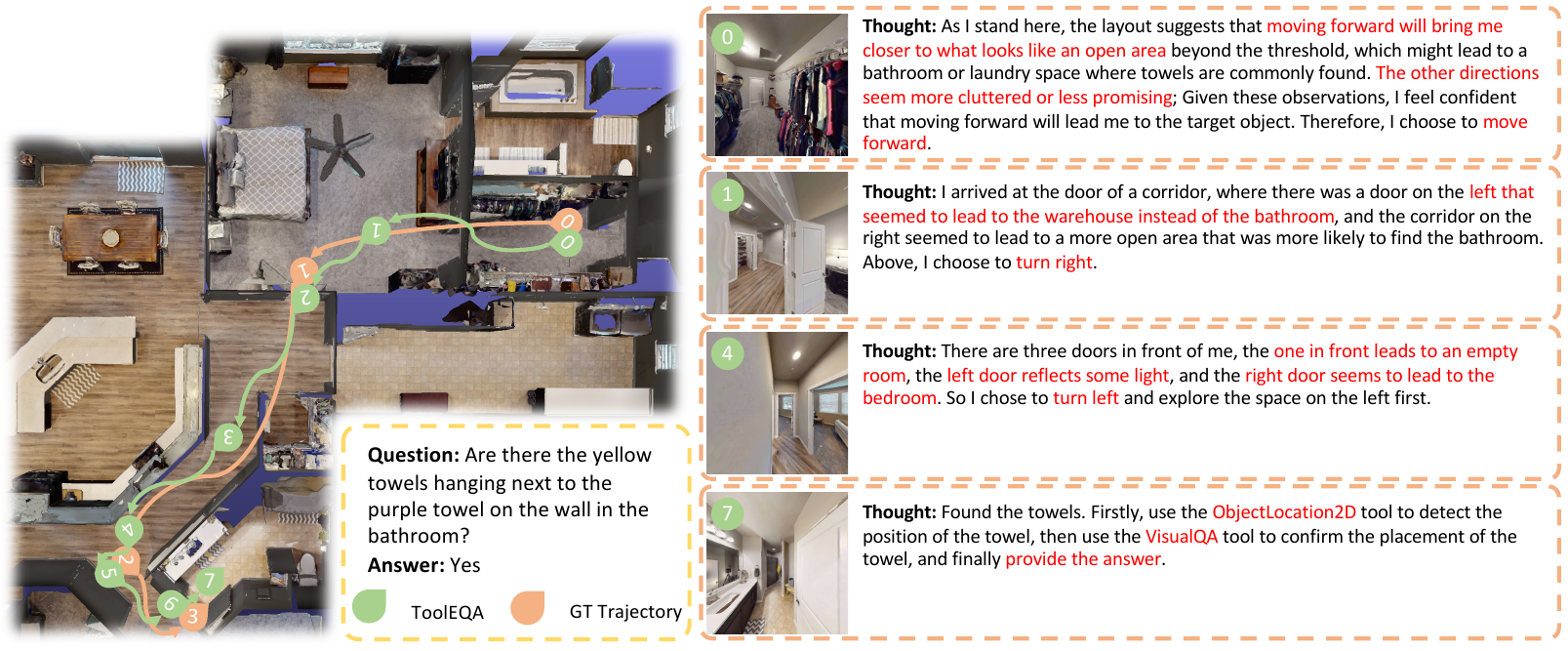}
    \vspace{-2em}
    \caption{
    Illustration of how explicit reasoning guides efficient exploration, enabling ToolEQA to answer questions faster and more accurately.
    }
    \label{fig:thought}
    \vspace{-1em}
\end{figure}
\begin{figure}[t]
    \centering
    \includegraphics[width=\linewidth]{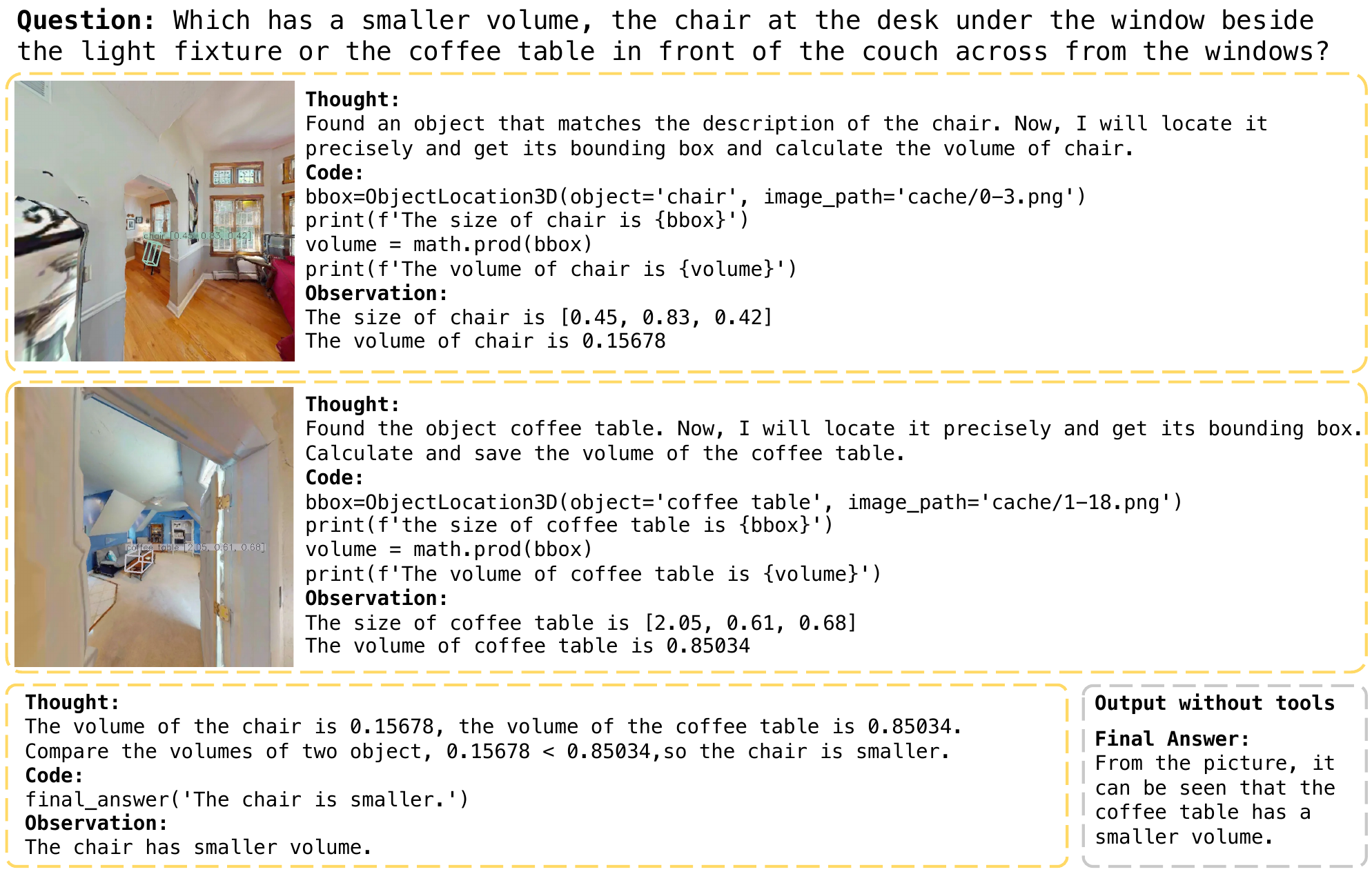}
    \vspace{-2em}
    \caption{
    Demonstration that the visual tools outperforms direct VLM inference without tools by accurately localizing, and comparing object volume.
    }
    \label{fig:tool-usage}
    \vspace{-1.5em}
\end{figure}

\vspace{-1em}
\subsection{Qualitative results}
\vspace{-0.5em}

As shown in Figure~\ref{fig:thought}, the decision-making process integrates spatial layout, functional space, and environmental cues to guide navigation toward the target object (towels). Each step is supported by clear reasoning, such as moving forward to approach potential bathroom space, turning right to explore a promising corridor, or turning left after excluding non-target rooms. This information-driven and reasoning-based decision paradigm ensures that the generated exploration trajectory maintains a high degree of proximity to the ground-truth trajectory, effectively validating the rationality and effectiveness of the decision-making framework in target-oriented spatial exploration tasks.

Figure~\ref{fig:tool-usage} highlights the clear advantage of tool-driven reasoning over direct Visual-Language Model (VLM) inference. Without tools, VLM often fails to localize objects precisely or distinguish fine-grained attributes such as size. In contrast, by integrating these specialized tools, our method obtained critical, fine-grained information (precise object localization, clutter-free cropping, and accurate size identification) that cannot be reliably captured by direct VLM inference on unprocessed images. This structured tool usage ensured the final comparison concluding that ``The chair has smaller volume'' was grounded in objective data, ultimately achieving a more accuracy response than would be possible with VLM alone.

\vspace{-1em}
\section{Conclusion}
\vspace{-1em}

In this work, we introduced \textbf{ToolEQA}, an embodied question answering agent that integrates explicit multi-step reasoning with tool usage to guide exploration and information acquisition. By coupling dynamic reasoning with executable tools, ToolEQA enables more efficient exploration paths and more reliable utilization of gathered observations. To support training, we proposed a scalable EQA data generation pipeline and constructed \textbf{EQA-RT}, a dataset of 18K automatically generated tasks with validated reasoning trajectories. Comprehensive experiments on HM-EQA, OpenEQA, ExpressBench and EQA-RT demonstrate that ToolEQA achieves significant improvements in both accuracy and efficiency over prior methods. 
These results highlight the importance of explicit multi-step reasoning and tool-usage in EQA agents, and suggest promising directions for developing more generalizable and interpretable frameworks for complex embodied AI tasks.

\newpage

\section{Ethics Statement}
This research adheres to the ICLR Code of Ethics. We confirm that all aspects of the study were conducted with the highest ethical standards. The work does not involve human subjects, and no personally identifiable or sensitive data were used. All datasets (HM3D, HM-EQA, OpenEQA, ExpressBench and EQA-RT) utilized were publicly available and comply with privacy regulations. We have taken steps to ensure that the models and algorithms developed are fair and free from bias. No conflicts of interest or external sponsorship influenced this work. Additionally, the findings are presented honestly, with all methodologies thoroughly validated for accuracy and reproducibility. We believe this research aligns with the ethical guidelines set forth by ICLR and contributes to the integrity and transparency of the scientific community.

\section{Reproducibility Statement}
To ensure the reproducibility of our results, we provide an anonymous code repository \url{https://anonymous.4open.science/r/ReactEQA-DDE3} that includes the complete implementation of data generation, data validation, and the ToolEQA framework. The repository also contains a comprehensive \texttt{README.md} file that details the installation requirements, dataset preparation, and step-by-step execution instructions. By following these guidelines, all experiments and results reported in this paper can be reproduced.

\bibliography{iclr2026_conference}

\begin{thebibliography}{29}
\providecommand{\natexlab}[1]{#1}
\providecommand{\url}[1]{\texttt{#1}}
\expandafter\ifx\csname urlstyle\endcsname\relax
  \providecommand{\doi}[1]{doi: #1}\else
  \providecommand{\doi}{doi: \begingroup \urlstyle{rm}\Url}\fi

\bibitem[Ahn et~al.(2022)Ahn, Brohan, Brown, Chebotar, Cortes, David, Finn, Fu, Gopalakrishnan, Hausman, et~al.]{ahn2022can}
Michael Ahn, Anthony Brohan, Noah Brown, Yevgen Chebotar, Omar Cortes, Byron David, Chelsea Finn, Chuyuan Fu, Keerthana Gopalakrishnan, Karol Hausman, et~al.
\newblock Do as i can, not as i say: Grounding language in robotic affordances.
\newblock \emph{arXiv preprint arXiv:2204.01691}, 2022.

\bibitem[Cangea et~al.(2019)Cangea, Belilovsky, Li{\`o}, and Courville]{cangea2019videonavqa}
C{\u{a}}t{\u{a}}lina Cangea, Eugene Belilovsky, Pietro Li{\`o}, and Aaron Courville.
\newblock Videonavqa: Bridging the gap between visual and embodied question answering.
\newblock \emph{arXiv preprint arXiv:1908.04950}, 2019.

\bibitem[Chen et~al.(2024)Chen, Qin, Zhang, Chen, Xu, and Che]{chen2024m}
Qiguang Chen, Libo Qin, Jin Zhang, Zhi Chen, Xiao Xu, and Wanxiang Che.
\newblock M3cot: A novel benchmark for multi-domain multi-step multi-modal chain-of-thought.
\newblock \emph{arXiv preprint arXiv:2405.16473}, 2024.

\bibitem[Cheng et~al.(2024)Cheng, Li, Sun, Min, Bedi, and Bera]{cheng2024efficienteqa}
Kai Cheng, Zhengyuan Li, Xingpeng Sun, Byung-Cheol Min, Amrit~Singh Bedi, and Aniket Bera.
\newblock Efficienteqa: An efficient approach for open vocabulary embodied question answering.
\newblock \emph{arXiv preprint arXiv:2410.20263}, 2024.

\bibitem[Das et~al.(2018{\natexlab{a}})Das, Datta, Gkioxari, Lee, Parikh, and Batra]{das2018embodied}
Abhishek Das, Samyak Datta, Georgia Gkioxari, Stefan Lee, Devi Parikh, and Dhruv Batra.
\newblock Embodied question answering.
\newblock In \emph{Proceedings of the IEEE conference on computer vision and pattern recognition}, pp.\  1--10, 2018{\natexlab{a}}.

\bibitem[Das et~al.(2018{\natexlab{b}})Das, Gkioxari, Lee, Parikh, and Batra]{das2018neural}
Abhishek Das, Georgia Gkioxari, Stefan Lee, Devi Parikh, and Dhruv Batra.
\newblock Neural modular control for embodied question answering.
\newblock In \emph{Conference on robot learning}, pp.\  53--62. PMLR, 2018{\natexlab{b}}.

\bibitem[Gao et~al.(2024)Gao, Zhang, Li, Ma, Yuan, Fan, Wu, Jia, Zhu, and Li]{gao2024multi}
Zhi Gao, Bofei Zhang, Pengxiang Li, Xiaojian Ma, Tao Yuan, Yue Fan, Yuwei Wu, Yunde Jia, Song-Chun Zhu, and Qing Li.
\newblock Multi-modal agent tuning: Building a vlm-driven agent for efficient tool usage.
\newblock \emph{arXiv preprint arXiv:2412.15606}, 2024.

\bibitem[Gordon et~al.(2018)Gordon, Kembhavi, Rastegari, Redmon, Fox, and Farhadi]{gordon2018iqa}
Daniel Gordon, Aniruddha Kembhavi, Mohammad Rastegari, Joseph Redmon, Dieter Fox, and Ali Farhadi.
\newblock Iqa: Visual question answering in interactive environments.
\newblock In \emph{Proceedings of the IEEE conference on computer vision and pattern recognition}, pp.\  4089--4098, 2018.

\bibitem[Hu et~al.(2022)Hu, Shen, Wallis, Allen-Zhu, Li, Wang, Wang, Chen, et~al.]{hu2022lora}
Edward~J Hu, Yelong Shen, Phillip Wallis, Zeyuan Allen-Zhu, Yuanzhi Li, Shean Wang, Lu~Wang, Weizhu Chen, et~al.
\newblock Lora: Low-rank adaptation of large language models.
\newblock \emph{ICLR}, 1\penalty0 (2):\penalty0 3, 2022.

\bibitem[Jiang et~al.(2025)Jiang, Liu, Chen, Luo, Chen, Pan, Li, and Lin]{jiang2025beyond}
Kaixuan Jiang, Yang Liu, Weixing Chen, Jingzhou Luo, Ziliang Chen, Ling Pan, Guanbin Li, and Liang Lin.
\newblock Beyond the destination: A novel benchmark for exploration-aware embodied question answering.
\newblock \emph{arXiv preprint arXiv:2503.11117}, 2025.

\bibitem[Jin et~al.(2024)Jin, Yu, Shu, Zhao, Hua, Meng, Zhang, and Du]{jin2024impact}
Mingyu Jin, Qinkai Yu, Dong Shu, Haiyan Zhao, Wenyue Hua, Yanda Meng, Yongfeng Zhang, and Mengnan Du.
\newblock The impact of reasoning step length on large language models.
\newblock \emph{arXiv preprint arXiv:2401.04925}, 2024.

\bibitem[Li et~al.(2025)Li, Li, Dong, Zhang, Zhang, Liu, Wang, Tang, and Liu]{li2025adaptive}
Wenjun Li, Dexun Li, Kuicai Dong, Cong Zhang, Hao Zhang, Weiwen Liu, Yasheng Wang, Ruiming Tang, and Yong Liu.
\newblock Adaptive tool use in large language models with meta-cognition trigger.
\newblock \emph{arXiv preprint arXiv:2502.12961}, 2025.

\bibitem[Li et~al.(2024)Li, Luo, Zhang, Qiu, Huang, and Wei]{li2024vocot}
Zejun Li, Ruipu Luo, Jiwen Zhang, Minghui Qiu, Xuanjing Huang, and Zhongyu Wei.
\newblock Vocot: Unleashing visually grounded multi-step reasoning in large multi-modal models.
\newblock \emph{arXiv preprint arXiv:2405.16919}, 2024.

\bibitem[Liu et~al.(2024)Liu, Hoang, Zhang, Zhu, Lan, Tan, Yao, Liu, Feng, RN, et~al.]{liu2024apigen}
Zuxin Liu, Thai Hoang, Jianguo Zhang, Ming Zhu, Tian Lan, Juntao Tan, Weiran Yao, Zhiwei Liu, Yihao Feng, Rithesh RN, et~al.
\newblock Apigen: Automated pipeline for generating verifiable and diverse function-calling datasets.
\newblock \emph{Advances in Neural Information Processing Systems}, 37:\penalty0 54463--54482, 2024.

\bibitem[Majumdar et~al.(2024)Majumdar, Ajay, Zhang, Putta, Yenamandra, Henaff, Silwal, Mcvay, Maksymets, Arnaud, et~al.]{majumdar2024openeqa}
Arjun Majumdar, Anurag Ajay, Xiaohan Zhang, Pranav Putta, Sriram Yenamandra, Mikael Henaff, Sneha Silwal, Paul Mcvay, Oleksandr Maksymets, Sergio Arnaud, et~al.
\newblock Openeqa: Embodied question answering in the era of foundation models.
\newblock In \emph{Proceedings of the IEEE/CVF conference on computer vision and pattern recognition}, pp.\  16488--16498, 2024.

\bibitem[OpenAI(2024{\natexlab{a}})]{gpt4o}
OpenAI.
\newblock Hello gpt-4o.
\newblock 2024{\natexlab{a}}.
\newblock URL \url{https://openai.com/index/hello-gpt-4o⁠}.

\bibitem[OpenAI(2024{\natexlab{b}})]{openai2024o1}
OpenAI.
\newblock Introducing openai o1-preview, 2024{\natexlab{b}}.
\newblock URL \url{https://openai.com/index/introducing-openai-o1-preview/}.

\bibitem[Ramakrishnan et~al.(2021)Ramakrishnan, Gokaslan, Wijmans, Maksymets, Clegg, Turner, Undersander, Galuba, Westbury, Chang, et~al.]{ramakrishnan2021habitat}
Santhosh~K Ramakrishnan, Aaron Gokaslan, Erik Wijmans, Oleksandr Maksymets, Alex Clegg, John Turner, Eric Undersander, Wojciech Galuba, Andrew Westbury, Angel~X Chang, et~al.
\newblock Habitat-matterport 3d dataset (hm3d): 1000 large-scale 3d environments for embodied ai.
\newblock \emph{arXiv preprint arXiv:2109.08238}, 2021.

\bibitem[Ranaldi et~al.(2024)Ranaldi, Pucci, Haddow, and Birch]{ranaldi2024empowering}
Leonardo Ranaldi, Giulia Pucci, Barry Haddow, and Alexandra Birch.
\newblock Empowering multi-step reasoning across languages via program-aided language models.
\newblock In \emph{Proceedings of the 2024 Conference on Empirical Methods in Natural Language Processing}, pp.\  12171--12187, 2024.

\bibitem[Ren et~al.(2024{\natexlab{a}})Ren, Clark, Dixit, Itkina, Majumdar, and Sadigh]{ren2024explore}
Allen~Z Ren, Jaden Clark, Anushri Dixit, Masha Itkina, Anirudha Majumdar, and Dorsa Sadigh.
\newblock Explore until confident: Efficient exploration for embodied question answering.
\newblock \emph{arXiv preprint arXiv:2403.15941}, 2024{\natexlab{a}}.

\bibitem[Ren et~al.(2024{\natexlab{b}})Ren, Jiang, Liu, Zeng, Liu, Gao, Huang, Ma, Jiang, Chen, et~al.]{ren2024grounding}
Tianhe Ren, Qing Jiang, Shilong Liu, Zhaoyang Zeng, Wenlong Liu, Han Gao, Hongjie Huang, Zhengyu Ma, Xiaoke Jiang, Yihao Chen, et~al.
\newblock Grounding dino 1.5: Advance the" edge" of open-set object detection.
\newblock \emph{arXiv preprint arXiv:2405.10300}, 2024{\natexlab{b}}.

\bibitem[Saxena et~al.(2024)Saxena, Buchanan, Paxton, Chen, Vaskevicius, Palmieri, Francis, and Kroemer]{saxena2024grapheqa}
Saumya Saxena, Blake Buchanan, Chris Paxton, Bingqing Chen, Narunas Vaskevicius, Luigi Palmieri, Jonathan Francis, and Oliver Kroemer.
\newblock Grapheqa: Using 3d semantic scene graphs for real-time embodied question answering.
\newblock \emph{arXiv preprint arXiv:2412.14480}, 2024.

\bibitem[Schick et~al.(2023)Schick, Dwivedi-Yu, Dess{\`\i}, Raileanu, Lomeli, Hambro, Zettlemoyer, Cancedda, and Scialom]{schick2023toolformer}
Timo Schick, Jane Dwivedi-Yu, Roberto Dess{\`\i}, Roberta Raileanu, Maria Lomeli, Eric Hambro, Luke Zettlemoyer, Nicola Cancedda, and Thomas Scialom.
\newblock Toolformer: Language models can teach themselves to use tools.
\newblock \emph{Advances in Neural Information Processing Systems}, 36:\penalty0 68539--68551, 2023.

\bibitem[Wang et~al.(2024)Wang, Bai, Tan, Wang, Fan, Bai, Chen, Liu, Wang, Ge, et~al.]{wang2024qwen2}
Peng Wang, Shuai Bai, Sinan Tan, Shijie Wang, Zhihao Fan, Jinze Bai, Keqin Chen, Xuejing Liu, Jialin Wang, Wenbin Ge, et~al.
\newblock Qwen2-vl: Enhancing vision-language model's perception of the world at any resolution.
\newblock \emph{arXiv preprint arXiv:2409.12191}, 2024.

\bibitem[Yao et~al.(2025)Yao, Huang, Qiu, Chen, Liu, Zhang, Zeng, Zhang, Zhang, Song, et~al.]{yao2025mmreason}
Huanjin Yao, Jiaxing Huang, Yawen Qiu, Michael~K Chen, Wenzheng Liu, Wei Zhang, Wenjie Zeng, Xikun Zhang, Jingyi Zhang, Yuxin Song, et~al.
\newblock Mmreason: An open-ended multi-modal multi-step reasoning benchmark for mllms toward agi.
\newblock \emph{arXiv preprint arXiv:2506.23563}, 2025.

\bibitem[Yao et~al.(2023)Yao, Zhao, Yu, Du, Shafran, Narasimhan, and Cao]{yao2023react}
Shunyu Yao, Jeffrey Zhao, Dian Yu, Nan Du, Izhak Shafran, Karthik Narasimhan, and Yuan Cao.
\newblock React: Synergizing reasoning and acting in language models.
\newblock In \emph{International Conference on Learning Representations (ICLR)}, 2023.

\bibitem[Yu et~al.(2019)Yu, Chen, Gkioxari, Bansal, Berg, and Batra]{yu2019multi}
Licheng Yu, Xinlei Chen, Georgia Gkioxari, Mohit Bansal, Tamara~L Berg, and Dhruv Batra.
\newblock Multi-target embodied question answering.
\newblock In \emph{Proceedings of the IEEE/CVF Conference on Computer Vision and Pattern Recognition}, pp.\  6309--6318, 2019.

\bibitem[Zhai et~al.(2025)Zhai, Gao, Wu, and Jia]{zhai2025memory}
Mingliang Zhai, Zhi Gao, Yuwei Wu, and Yunde Jia.
\newblock Memory-centric embodied question answer.
\newblock \emph{arXiv preprint arXiv:2505.13948}, 2025.

\bibitem[Ziliotto et~al.(2025)Ziliotto, Campari, Serafini, and Ballan]{ziliotto2025tango}
Filippo Ziliotto, Tommaso Campari, Luciano Serafini, and Lamberto Ballan.
\newblock Tango: training-free embodied ai agents for open-world tasks.
\newblock In \emph{Proceedings of the Computer Vision and Pattern Recognition Conference}, pp.\  24603--24613, 2025.

\end{thebibliography}
\bibliographystyle{iclr2026_conference}

\newpage
\appendix
\section{Appendix}

\subsection{Data Statistic}
\label{sec:data_statistic}
\begin{figure}[htbp]
    \centering
    \begin{subfigure}{0.3\textwidth}
        \centering
        \includegraphics[width=\linewidth]{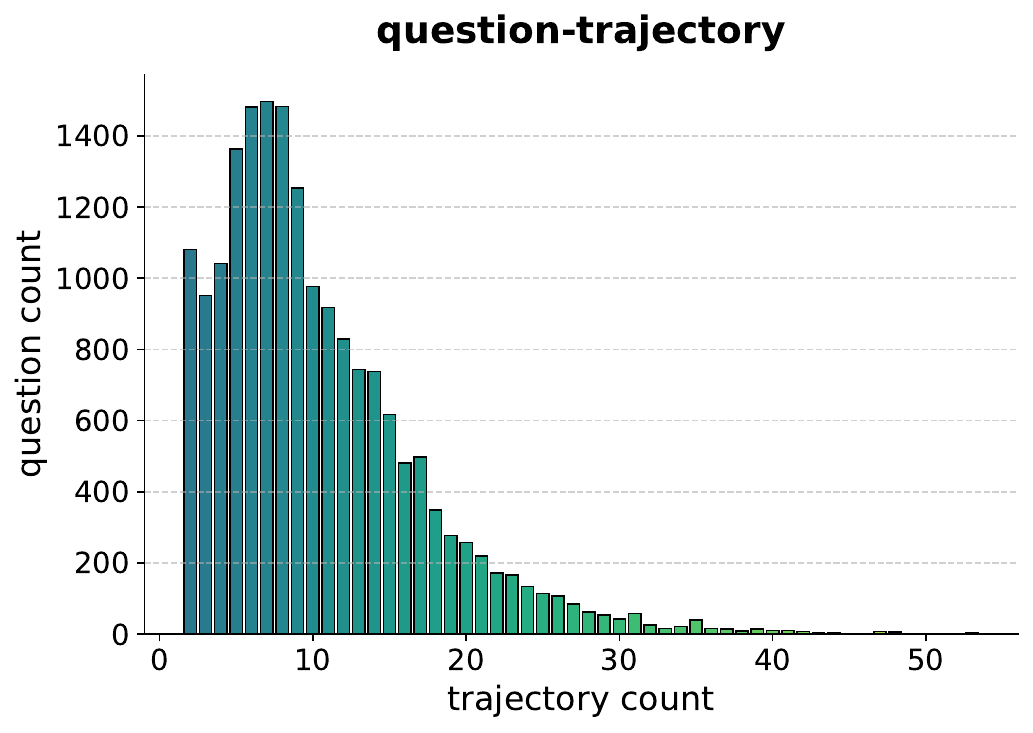}
        \caption{Statistics on the number of questions with different steps counts.}
    \end{subfigure}
    \hfill
    \begin{subfigure}{0.3\textwidth}
        \centering
        \includegraphics[width=\linewidth]{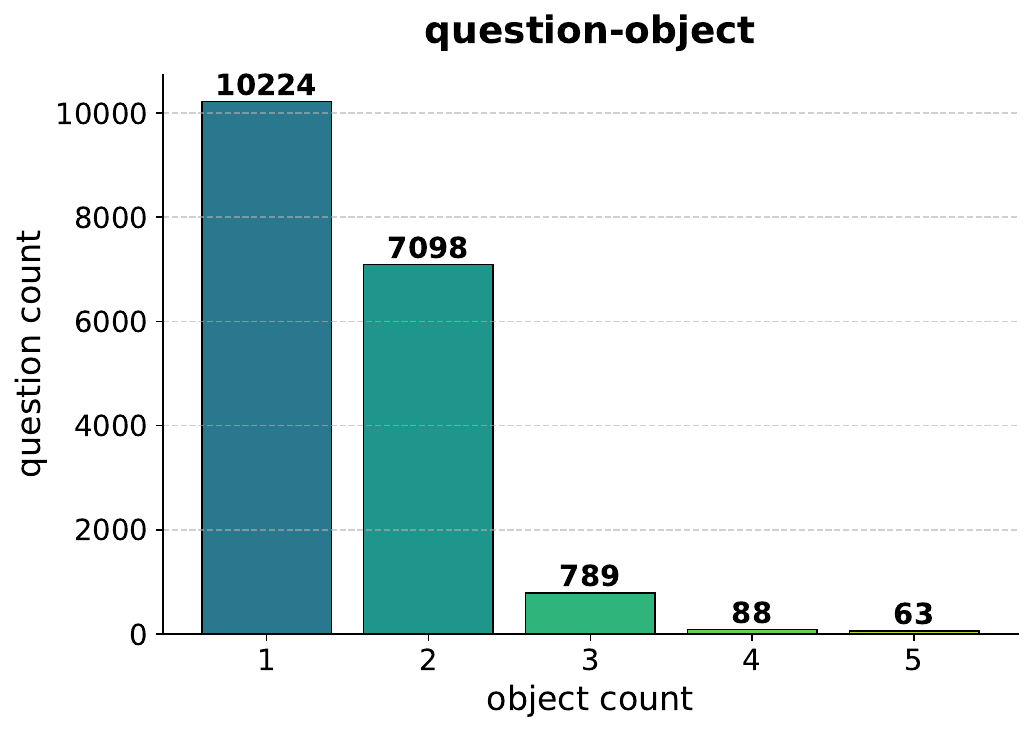}
        \caption{Statistics on the number of questions with different object counts.}
    \end{subfigure}
    \hfill
    \begin{subfigure}{0.3\textwidth}
        \centering
        \includegraphics[width=\linewidth]{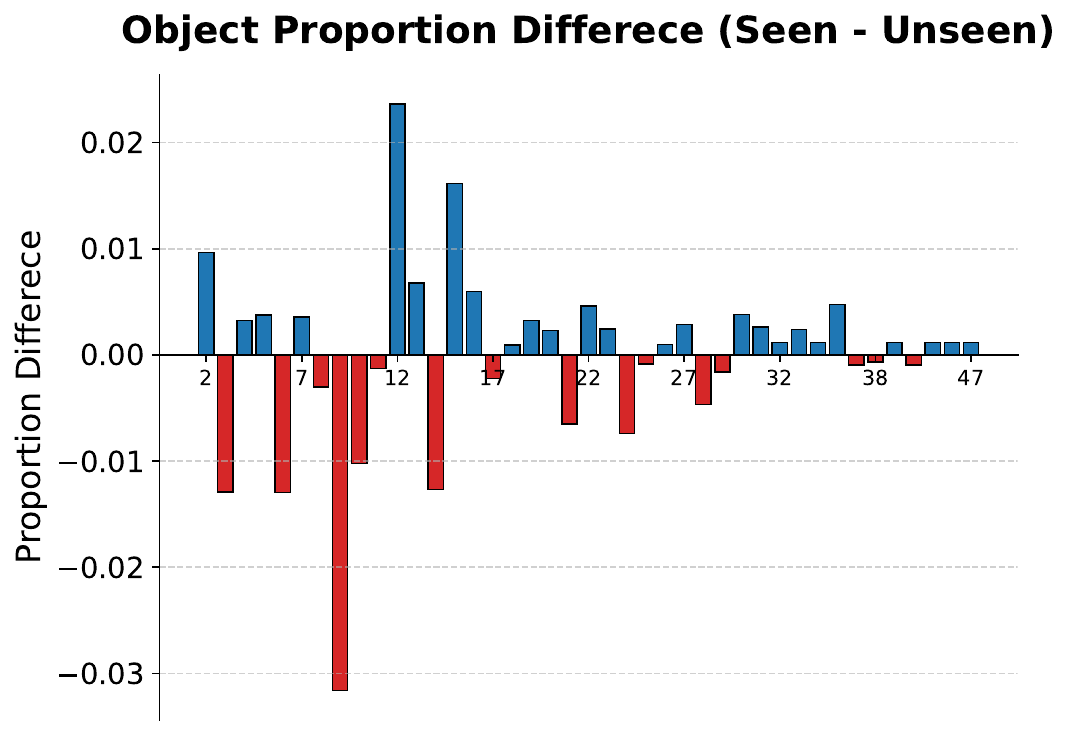}
        \caption{The difference in the ratio of the average number of objects between the EQA-RT-Seen and the EQA-RT-Unseen.}
    \end{subfigure}
    \caption{Dataset statistic about steps count, objects count and the difference between two test set.}
    \label{fig:data_statistic}
\end{figure}

\begin{table}[h]
    \centering
    \caption{Statistic about average exploration steps, the average number of tools used per task, and the average exploration length.}
    \begin{tabular}{c|ccc}
    \toprule
         & Step & Tool & Length (m) \\
    \midrule
        EQA-RT-Train &  12.74 & 12.39 & 13.13 \\
        EQA-RT-Seen & 12.56 & 12.20 & 12.71 \\
        EQA-RT-Unseen & 12.13 & 11.76 & 12.38 \\
        EQA-RT & 12.69 & 12.35 & 13.07 \\
    \bottomrule
    \end{tabular}
    \label{tab:step_tool_length}
\end{table}

As shown in Figure~\ref{fig:data_statistic}, the dataset exhibits a pronounced long-tail distribution in both the average exploration steps and the number of related objects per question. Most questions require around ten exploration steps; among them, 10,224 involve a single target, 7,098 involve two targets, and 940 involve three or more objects. Figure~\ref{fig:data_statistic}(c) further compares EQA-RT-Seen and EQA-RT-Unseen in terms of the average number of objects per question, revealing that EQA-RT-Seen involves more objects. Since object count reflects task difficulty, this suggests that tasks in EQA-RT-Seen are more challenging.

In addition, as shown in Table~\ref{tab:step_tool_length}, we report the exact values of the average exploration steps, the average number of tools used per task, and the average exploration length across different sets. By comparing the statistics of EQA-RT-Seen and EQA-RT-Unseen, it can also be inferred that the tasks in EQA-RT-Seen are more challenging.

\subsection{Metric Details}
\label{sec:details_of_metric}
To comprehensively evaluate the effectiveness of our approach, we use $recall@D$, $e_{path}@D$ and success rate as metrics.

The success rate is divided into two parts, for multi choices tasks, we calculate average accuracy between the output of model and ground truth answer; for open vocabulary, we prompt LLM to obtain the semantic similarity $\sigma_i\in\{0,1,2,3,4,5\}$ between the output of model and ground truth answer, and then calculate $\textit{LLM-Match Score}=\frac{1}{N}\sum_{i=1}^N\frac{\sigma_i}{5}\times 100\%$.

The $recall@D$ is used to evaluate whether objects related to the problem were found during the exploration process. 
So we first define $n$ be the number of objects and $T$ the number of camera steps. 
At step $t$, the camera position is $\mathbf{p}_t \in \mathbb{R}^3$ and its yaw angle (around the $y$-axis) is $\theta_t$. 
The forward unit vector of the camera is $\mathbf{f}_t = (\sin\theta_t,\,0,\,-\cos\theta_t)$. 
The position of object $j$ is $\mathbf{o}_j \in \mathbb{R}^3$, and the distance from the camera to the object is $d_{j,t} = \|\mathbf{o}_j - \mathbf{p}_t\|$. 
The $recall@D$ can be formalized as 
$$
recall@D = \frac{1}{n}\sum_{j=1}^{n}\;
\max_{t}\left\{
\Big(1 - \tfrac{d_{j,t}}{D}\Big)\;
\mathbf{1}\!\left[
d_{j,t}\le D,\;
\frac{\mathbf{f}_t\cdot(\mathbf{o}_j-\mathbf{p}_t)}{\|\,\mathbf{o}_j-\mathbf{p}_t\,\|}
\;\ge\; \cos\!\left(\tfrac{\mathrm{FOV}}{2}\right)
\right]
\right\},
$$
where $d_{j,t}=||\mathbf{o}_j-\mathbf{p}_t||$, $\mathbf{f}_t=(\sin{{\theta_t},0,-\cos{\theta_t}})$, $\mathbf{1[\cdot]}$ is indicator function.

The $e_{path}@D$ is an indicator that combines success rate, recall, and exploration path length. 
The specific calculation process is as follows 
$$
e_{path}@D = \frac{1}{N}\sum_{j=1}^N(\text{success rate})\times recall@D \times \exp(\frac{l_i}{\max(p_i,l_i)}),
$$
where $l_i$ is length of the shortest path, $p_i$ is length of the exploring path.

\subsection{Tools Description}
\label{sec:tools_desc}
\begin{table}[h]
    \centering
    \caption{The description of tools.}
    \begin{tabular}{c|p{10cm}}
    \toprule
        Tool & Description \\
    \midrule
        GoNextPoint & The agent conitnue explore next point and obtain next observation (rgb image). \\
    \midrule
        ObjectLocation2D & A tool that can localize objects in given images, outputing the bounding boxes of the objects. \\
    \midrule
        ObjectLocation3D & Localize 3D objects in the scene and return their 3D bounding boxes and center coordinates. \\
    \midrule
        ObjectCrop & Given the bounding boxes of objects, crop and save the relevant objects from the image. \\
    \midrule
        SegmentInstance & A tool that can do instance segmentation on the given image. \\
    \midrule
        VisualQA & A tool that can answer questions about attached images. \\
    \midrule
        FinalAnswer & Provides a final answer to the given problem. \\
    \bottomrule
    \end{tabular}
    \label{tab:tool_desc}
\end{table}

As shown in Table~\ref{tab:tool_desc}, we present all the tools used and their corresponding descriptions.

\subsection{Efficiency Analysis}

\begin{table}[h]
    \centering
    \caption{Efficiency Analysis about time consumption (Time), LLM Token usage (Token) and GPU Memory usage (Memory).}
    \begin{tabular}{c|ccc}
    \toprule
         & Time (s) & Token & Memory (G)\\
    \midrule
        Planner     & 0.52  & 118     & -     \\
        Controller  & 25.4 & 9512.7  & 40.5  \\
        Executor    & 5.54  & -       & -     \\
    \bottomrule
    \end{tabular}
    \label{tab:efficiency}
\end{table}

We conducted an efficiency analysis of ToolEQA, and Table~\ref{tab:efficiency} reports the time consumption, LLM token consumption, and memory usage of each module. It is worth noting that the planner is executed only once before the exploration begins, whereas the controller and executor run continuously throughout the entire exploration process.

\subsection{The Use of Large Language Models}
A large language model (ChatGPT, Deepseek-R1, Doubao) was employed during manuscript preparation. The model was used for grammar checking, sentence refinement, and improving the readability of the text. In addition, the model was consulted for assistance in drafting segments of project code, primarily for debugging and improving code efficiency. Within the ToolEQA framework, large language models were also utilized as the Controller, including GPT-4o and Qwen2.5VL-7B, to generate control decisions during exploration. All scientific ideas, analyses, and final implementations were designed and verified by the authors.

\end{document}